\documentclass[letterpaper]{article}
\usepackage{flairs}%aaai
\usepackage{times}
\usepackage{helvet}
\usepackage{courier}

\usepackage{pgfplots}
\pgfplotsset{compat=1.14, every non boxed x axis/.append style={x axis line style=-},
     every non boxed y axis/.append style={y axis line style=-}}
\usepgfplotslibrary{fillbetween}
\usepackage{amsmath}
\usepackage{booktabs}
\usepackage{multirow}
\usepackage{pifont}

\usepackage{nameref}

\usepackage[hyphens]{url}  % DO NOT CHANGE THIS

\begin{document}
% This file is an adoption of the style file for AAAI Press 
% proceedings, working notes, and technical reports.  This file is made 
% with minimal changes by explicit permission from AAAI.
%
\title{Query-Based Keyphrase Extraction from Long Documents}
\author{Martin Docekal, Pavel Smrz\\
Brno University of Technology\\
idocekal@fit.vutbr.cz, smrz@fit.vutbr.cz
}
\copyrighttext {Copyright \copyright\space 2022 by the authors.
All rights reserved.}
\maketitle

\begin{abstract}
Transformer-based architectures in natural language processing force input size limits that can be problematic when long documents need to be processed. This paper overcomes this issue for keyphrase extraction by chunking the long documents while keeping a global context as a query defining the topic for which relevant keyphrases should be extracted. The developed system employs a pre-trained BERT model and adapts it to estimate the probability that a given text span forms a keyphrase. We experimented using various context sizes on two popular datasets, Inspec and SemEval, and a large novel dataset. The presented results show that a shorter context with a query overcomes a longer one without the query on long documents.\footnote{The code is available at https://github.com/KNOT-FIT-BUT/QBEK.}
\end{abstract}

\clubpenalty=10000
\widowpenalty=10000
\displaywidowpenalty=10000

\section{Introduction}
% Keyword and keyphrase are terms with slightly varying meanings across fields they are explored in -- linguistics, computer science, information retrieval, library science, knowledge organization, and social sciences (see the discussion on the definitions and related terms in~\cite{keywordISKO}). 
Keyphrase refers to a short language expression describing the content of a longer text.
%If a keyphrase consists of a single word, it is called a keyword (we use these terms interchangeably).
Due to their concise form, keyphrases can be used for a quick familiarization with a document.
They also improve the findability of documents or passages within them. 
In the bibliographic records, keyphrase descriptors enable flexible indexing.
%Keyphrases can also play an important role in automatic text summarization~\cite{summ} (especially for long documents~\cite{neco}), ontology construction~\cite{onto}, recommender systems~\cite{recom}, as well as in many other sub-components of information technology. 

% In addition to their primary function, keyphases can be categorized along various dimensions. The notion of ``keyness'' is thoroughly discussed in~\cite{firoozeh2020keyword}. The typology of the agent -- human or computer-based -- plays a crucial role. Although it is seen as a dichotomy in some works~\cite{hjorland2011importance}, there is a wide range of middle cases. For example, \textit{\#hastags} provided by authors of social media posts can be interpreted as a kind of keywords. Similarly to that, the subject headings are added manually by librarians to bibliographic records corresponding to entire books but they can be used to characterize the content of individual passages or even sentences too. 

Whether a text span is a keyphrase depends on the context of that span because a keyphrase for a specific topic may not be a keyphrase for another topic. The presented work builds on the idea that the topic can be explicitly given as an input to the keyphrase extraction algorithm in the form of a query. We approximate such a query with a document's title in our experiments. We also investigate the influence of context size and document structure on the results.

\section{Related Work}
Traditional approaches to keyphrase extraction involve graph-based methods, such as TextRank \cite{textrank} and RAKE \cite{rake}.
Recently, many types of neural networks have been used for the task \cite{lin2019keyword,sahrawat2020keyphrase}. Most of the deep learning work assumes the existence of a title and an abstract of the document and extracts keyphrases from them because they struggle with longer inputs such as whole scientific articles \cite{kontoulis-etal-2021-keyphrase}. Some works try to overcome this limitation by first creating a document summary and then extracting keyphrases from it \cite{kontoulis-etal-2021-keyphrase,kb-rank}. Our research follows an alternative path, compensating for the limited context by a query specifying a topic.

\section{Model}
\begin{figure}
    \centering
    \includegraphics[width=8cm]{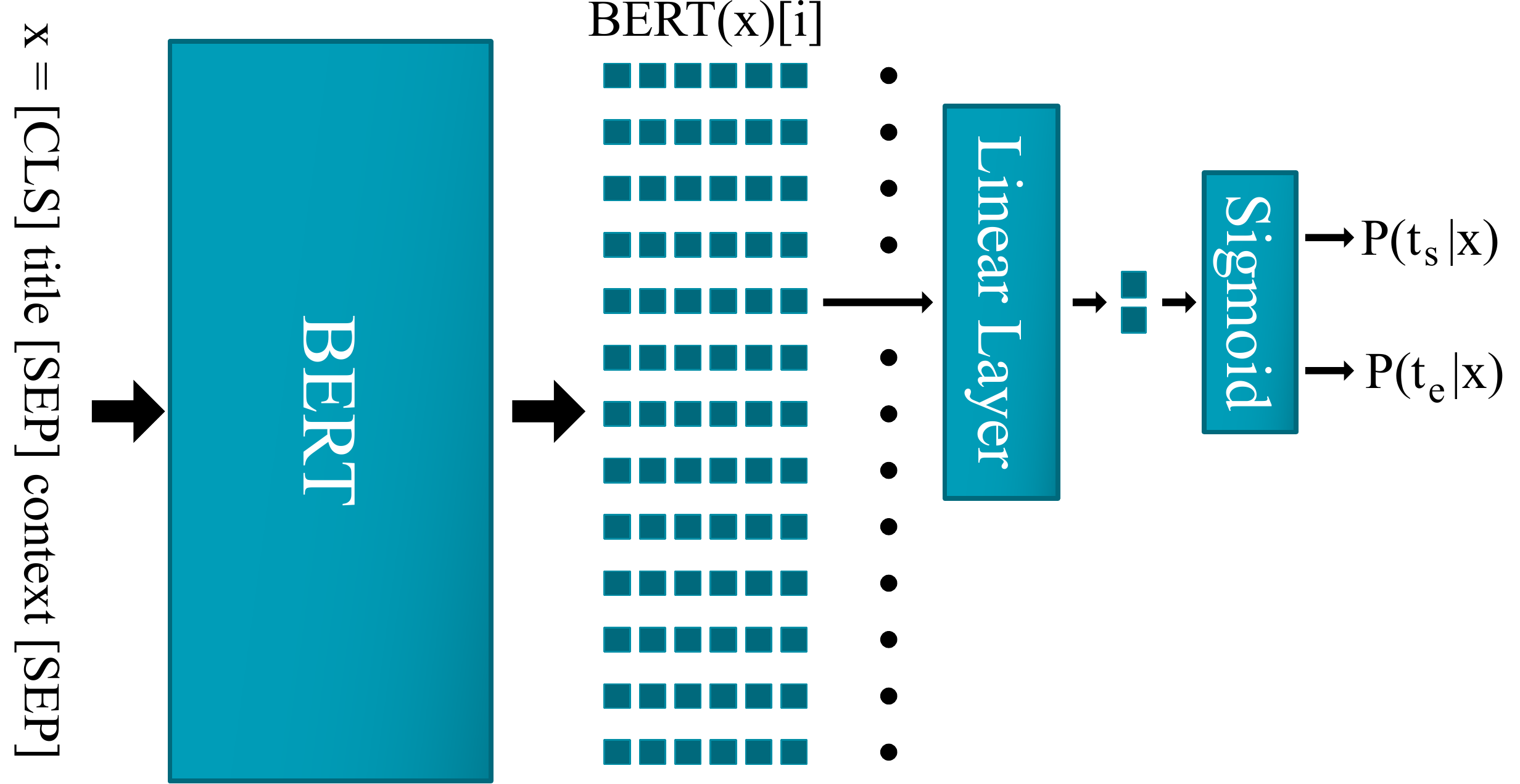}
    \caption{Illustration of our model with query at the input.}
    \label{fig:model}
\end{figure}

First, a document is split into parts (contexts), which are further processed independently. Then, the devised model estimates the probability that a given continuous text span forms a keyphrase. It looks for boundaries $t_s$ and $t_e$, corresponding to the text span's start and end, respectively. The inspiration for this approach comes from the task of reading comprehension, where a similar technique is used to search for potential answers to a question in an input text~\cite{BERT}. Formally, the model estimates probability:
\begin{equation}
    P(t_s,t_e| x) = P(t_s|x)P(t_e|x) \, ,
\end{equation}
where  $x$ is the input sequence. It assumes that the start and the end of a text span $<t_s, t_e>$ are independent.
The probabilities $P(t_s|x)$ and $P(t_e|x)$ are obtained in the following way:
\begin{equation}
    P(t_*|x)=\text{sigmoid}(w_{*}^T \text{BERT}(x)[*] + b) \, ,
\end{equation}
where $*$ stands for the \emph{start} and \emph{end} positions. Weights $w_{s}$ and $w_{e}$ are learnable vectors, $b$ is the scalar bias and $ \text{BERT}(x)[i]$ is BERT \cite{BERT} vector representation of a token from sequence $x$ on position $i$. See the model illustration in Figure \ref{fig:model}.

The task of predicting whether a given token is the start token of a span or the end token could be seen as binary classification with two classes \textit{start/end} and \textit{not start/not end}, respectively. The binary cross-entropy ($\operatorname{BCE}$) loss function is used for training in the following way:
\begin{equation}
    \operatorname{BCE}(v_s, g_s) + \operatorname{BCE}(v_e, g_e),
\end{equation}
where $v_s$ is a vector of probabilities that a token is the start token of a span, for each token in the input, and $v_e$ is vector computed analogously, but for the ends. The $g_s$ and $g_e$ are vectors of ground truth probabilities of starts or ends, respectively.

We work with two types of inputs. One consists of a text fragment such as a sentence, and the other uses a query (document title) and a text segment, separated by a special token. Various context sizes are explored in our experiments. The context size determines how big is the document part the model sees at once. Every context part of a document is processed independently. The final list of keyphrases is created by collecting keyphrase spans with their scores and selecting the top ones.

\section{Datasets}

Besides two standard datasets for keyphrase extraction, we created and used a novel dataset of long documents, referred to a Library, and we also prepared an unstructured version of the SemEval-2010 dataset. A comparison of the datasets is given in Table \ref{tab:datasets}.

\begin{table}[h]
    \centering
    \begin{tabular}{@{}rcccc@{}}
    \toprule
    \textbf{dataset} & \textbf{train} & \textbf{val.} & \textbf{test} & \begin{tabular}[c]{@{}c@{}}\textbf{sentences}\\\textbf{(train)}\end{tabular} \\ \midrule
    SemEval-2010     & 130            & 14                  & 100   & 66~428 (72\%)        \\
    \begin{tabular}[c]{@{}r@{}}Unstructured-\\SemEval-2010\end{tabular}     & 130            & 14                  & 100   & 45~346 (67\%)        \\
    Inspec           & 1~000           & 500                 & 500   & 5~894 (25\%)          \\
    Library          & 48~879          & 499                 & 499   & 298~217~589 (94\%)          \\ \bottomrule
    \end{tabular}
    \caption{The number of documents in each split along with the total number of sentences in a train set. The percentage in the sentences column is the proportion of sentences without keyphrases.}
    \label{tab:datasets}
\end{table}

We had to annotate the spans that represent given keyphrases in the text as the discussed datasets provide just a list of associated keyphrases with no information about their actual positions. The search was case insensitive and
the Porter stemmer was utilized for the SemEval and Hulth2003 (Inspec) datasets. For the Library dataset, as it is in Czech, the \textit{ MorphoDiTa} lemmatizer\footnote{http://hdl.handle.net/11858/00-097C-0000-0023-43CD-0} was used.

\subsubsection{SemEval-2010}
\cite{semeval-2010} consists of whole plain texts from scientific articles. The dataset provides keyphrases provided by authors and readers. As it is common practice \cite{semeval-2010,kontoulis-etal-2021-keyphrase}, we use a combination of both in our experiments. Our validation dataset was created by randomly choosing a subset of the train set. As the original data source does not explicitly provide the titles, which we need to formulate a query, we have manually extracted the title of each document from the plain text.

% The authors selected articles from multiple categories: Distributed Systems, Information Search and Retrieval, Distributed Artificial Intelligence – Multiagent Systems, and Social and Behavioral Sciences – Economics. The articles were obtained from the ACM Digital Library.

Documents in this dataset have a well-formed structure. They contain a title and abstract and are divided into sections introduced with a headline. As we want to investigate the influence of such structure on results, we have made an unstructured version of this dataset. We downloaded the original PDFs and used the GROBID\footnote{https://github.com/kermitt2/grobid} to get a structured XML version of them. We kept only the text from the document's main body while the parts such as title, abstract, authors, or section headlines were removed. Nevertheless, document keyphrase annotations remain the same. We call this dataset Unstructured-SemEval-2010. The name SemEval is used to name these two collectively.

\subsubsection{Inspec}
\cite{hulth2003} contains a set of title-abstract pairs collected from scientific articles. For each abstract, there are two sets of keyphrases — \textit{controlled}, which are restricted to the Inspec thesaurus, and \textit{uncontrolled} that can contain any suitable keyphrase. To be comparable with previous works \cite{hulth2003,kb-rank}, we used only the \textit{uncontrolled} set in our experiments. 

%Annotated keyphrases may or may not be presented in the corresponding text. So this dataset can be used for keyphrase extraction and also for keyphrase generation.

\subsubsection{Library}
is a newly created dataset that takes advantage of a large body of scanned documents, provided by Czech libraries, that were converted to text by OCR software. This way of getting the text is unreliable, so the dataset contains many errors on the word and character level. The dataset builds on the documents where the language was recognized as `Czech' by the OCR software.

%\footnote{Miss annotated documents were observed so the dataset may contain documents in other languages (mainly German and English)}.

All used documents in the original data source are represented by their significant content (the average number of characters per document is 529~276) and metadata. The metadata contains (not for all) keyphrases and document language annotations. We did not ask annotators to annotate each document. Instead, we selected metadata fields used by librarians as keyphrase annotations. So, our data and metadata come from the real-world environment of Czech libraries. We have filtered out all documents with less than five keyphrases.

Documents come from more than 290 categories. Various topics such as mathematics, psychology, belles lettres, music, and computer science are covered. Not all annotated keyphrases can be extracted from the text. Considering the lemmatization, the test set annotations contain about 53\% of extractive keyphrases. Bibliographic field Title (MARC~245\footnote{https://www.loc.gov/marc/bibliographic/bd245.html}) is used as the query. Note that the field may contain additional information to the title, such as authors.

\section{Experimental Setup}\label{sec:experimental_setup}
The implemented system builds on PyTorch\footnote{https://pytorch.org/} and PyTorch Lightning\footnote{https://www.pytorchlightning.ai/} tools. The BERT part of the model uses the implementation of BERT by \textit{Hugging Face}\footnote{https://huggingface.co/} and it is initialized with pre-trained weights of \textit{bert-base-multilingual-cased}. These weights are also optimized during fine-tuning.

The Adam optimiser with weight decay \cite{adamW} is used in all the experiments. The evaluation during training on the validation set is done every 4~000 optimization steps for the Library dataset and every 50 steps for Inspec (25 for whole abstracts with titles). For SemEval datasets, the number of steps differs among experiments. Early stopping with patience 3 is applied, so the training ends when the model stops improving. Batch size 128 is used for experiments with the Library dataset, and batch size~32 is used for Inspec and SemEval datasets. The learning rate 1E-06 is used for the experiments with SemEval datasets, while it is set to the value of 1E-05 for all other datasets. Inputs longer than a maximum input size are split into sequences of roughly the same size in a way that forbids splitting of keyphrase spans. In edge cases (split is not possible), the input is truncated. No predicted span is longer than 6 tokens.

The official script for SemEval-2010 is used for evaluation. However, the normalization of keyphrases is different for the Library dataset as we have used the mentioned Czech lemmatizer instead of the original stemmer. We use the F1 over the top five (F1@5) candidates for the Library dataset and over the top ten (F1@10) for the rest.

\section{Experiments}
The performed experiments investigate the influence of queries on four different datasets, the output quality with various context sizes, and the impact of the document structure.

\begin{figure}
    \centering
    \begin{tikzpicture}
    \begin{axis}[
            %smooth,
            thick,
            legend columns=2,
            legend style={
                at={(0.4,1.05)}, 
                legend cell align={left},  
                anchor=south
            },
            xlabel={input size [sentences]},
            ylabel=F1@10,
            width=1.0\columnwidth,
            height=0.7\columnwidth,
            xtick=data,
            xmajorgrids,
            ymajorgrids,
        ]
        \addplot+[red!50,mark options={fill=red!50}] table [x=sentences, y=sentences only, col sep=comma] {data/semeval.csv};
        \addplot+[red!100,mark options={fill=red}] table [x=sentences, y=sentences + title, col sep=comma] {data/semeval.csv};
        \addplot+[blue!50,mark=triangle, mark options={fill=blue}] table [x=sentences, y=unstructured - sentences only, col sep=comma] {data/semeval.csv};
        \addplot+[blue!100, mark options={fill=blue}] table [x=sentences, y=unstructured - sentences + title, col sep=comma] {data/semeval.csv};
        
        \addplot [name path=l unstructured - sentences only, fill=none, draw=none] table [
            x=sentences, y expr=\thisrow{unstructured - sentences only} - \thisrow{CI95 unstructured - sentences only}, col sep=comma]{data/semeval.csv};
        \addplot [name path=u unstructured - sentences only, fill=none, draw=none] table [
            x=sentences, y expr=\thisrow{unstructured - sentences only} + \thisrow{CI95 unstructured - sentences only}, col sep=comma]{data/semeval.csv};
        \addplot[blue!40, opacity=0.2] fill between[of=l unstructured - sentences only and u unstructured - sentences only];
        
        \addplot [name path=l sentences title, fill=none, draw=none] table [
            x=sentences, y expr=\thisrow{sentences + title} - \thisrow{CI95 sentences + title}, col sep=comma]{data/semeval.csv};
        \addplot [name path=u sentences title, fill=none, draw=none] table [
            x=sentences, y expr=\thisrow{sentences + title} + \thisrow{CI95 sentences + title}, col sep=comma]{data/semeval.csv};
        \addplot[red!40, opacity=0.2] fill between[of=l sentences title and u sentences title];

        %\addplot table [x=test_input_size, y=16, col sep=comma] {data/extractive-reader-batch-sizes.csv};
        %\addplot+[mark=triangle*] table [x=test_input_size, y=32, col sep=comma] {data/extractive-reader-batch-sizes.csv};
        %\addplot table [x=test_input_size, y=64, col sep=comma] {data/extractive-reader-batch-sizes.csv};
        %\addplot+[color=orange,mark options={fill=orange}] table [x=test_input_size, y=128, col sep=comma] {data/extractive-reader-batch-sizes.csv};

        %\addplot[dashed,thick, samples=50, smooth,domain=0:6,black] coordinates {(43,0.4)(43,0.9)};
        \addlegendentry{sentences only}
        \addlegendentry{sentences + query}
        \addlegendentry{sentences only (unst)}
        \addlegendentry{sentences + query (unst)}
    \end{axis}%
\end{tikzpicture}%
    \caption{Results for SemEval-2010 and Unstructured-SemEval-2010 test set. The light red and blue areas are confidence intervals with a confidence level of 0.95. Each point corresponds to an average of five runs.}
    \label{fig:sem_eval_results}
\end{figure}
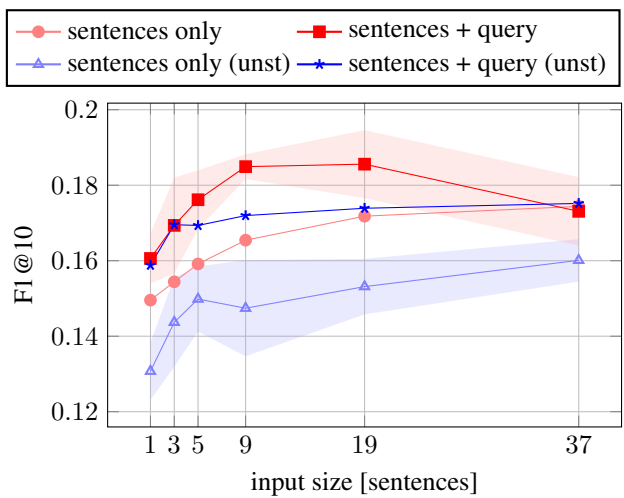

\begin{figure}
    \centering
    \begin{tikzpicture}
    \begin{axis}[
            %smooth,
            thick,
            legend columns=2,
            legend style={
                at={(0.5,1.05)}, 
                legend cell align={left},  
                anchor=south
            },
            xlabel={input size [sentences]},
            ylabel={too long inputs [\%]},
            ytick={0.0,25,50,75},
            width=1\columnwidth,
            height=0.40\columnwidth,
            xtick=data,
            xmajorgrids,
            ymajorgrids,
        ]
        \addplot table [x=sentences, y=saturation, col sep=comma] {data/semeval_2010_saturation.csv};
        \addplot table [x=sentences, y=with title saturation, col sep=comma] {data/semeval_2010_saturation.csv};
        
        \addlegendentry{sentences only}
        \addlegendentry{sentences + query}
        
    \end{axis}%
\end{tikzpicture}%
    \caption{The proportion of inputs longer than the maximum input size for SemEval-2010 train set.}
    \label{fig:sem_eval_saturation}
\end{figure}
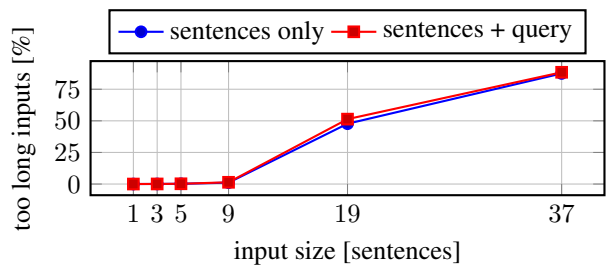

The first set of experiments is performed on long documents with a well-formed structure from the SemEval-2010 dataset and compares them with SemEval's unstructured version. Figure \ref{fig:sem_eval_results} shows that inputs with a query are better than those without a query, but the last point. For the structured input, it can be seen that from the point with 19 sentences, the performance of input with query stops with the fast growth. It correlates with Figure \ref{fig:sem_eval_saturation} showing the saturation of the model input. Notice that from 19 sentences, the input becomes more saturated, and the splitting strategy starts shrinking contexts.

It is not surprising that the nominal values are lower for unstructured inputs. On the other hand, it is clear that the query has a bigger influence on the unstructured version, especially for short context sizes, because the average absolute difference among results (with- and without a query) for each context size is 2.2\% compared to 1.29\% for the structured one. 

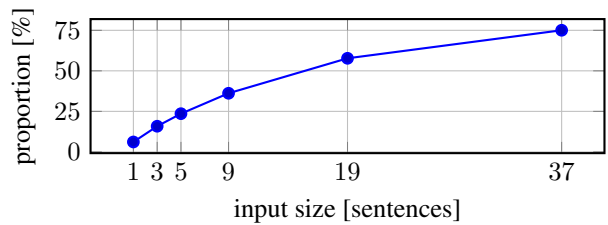
\begin{figure}
    \centering
    \begin{tikzpicture}
    \begin{axis}[
            %smooth,
            thick,
            xlabel={input size [sentences]},
            ylabel={proportion [\%]},
            ytick={0.0,25,50,75},
            width=1\columnwidth,
            height=0.4\columnwidth,
            xtick=data,
            xmajorgrids,
            ymajorgrids,
        ]
        \addplot table [x=sentences, y=proportion, col sep=comma] {data/semeval_proportion_of_headlines.csv};
        
    \end{axis}%
\end{tikzpicture}%
    \caption{The proportion of contexts containing at least one section headline as a substring for SemEval-2010 test set.}
    \label{fig:sem_eval_headlines_proportion}
\end{figure}

Looking at the curve of results with a query on an unstructured version, we assume that the model can exploit a document structure without explicitly tagging it with special tokens because additional context size above the three sentences is not much beneficial compared to the case with the document structure. This hypothesis is supported by the fact that the proportion of the context containing structured information grows with context size, as is demonstrated in Figure \ref{fig:sem_eval_headlines_proportion} showing the proportion of contexts containing a section headline.

\begin{figure}
    \centering
    \begin{tikzpicture}
    \begin{axis}[
            %smooth,
            thick,
            legend columns=2,
            legend style={
                at={(0.5,1.05)}, 
                legend cell align={left},  
                anchor=south
            },
            xlabel={input size [sentences]},
            ylabel=F1@5,
            width=1\columnwidth,
            height=0.5\columnwidth,
            xtick=data,
            xmajorgrids,
            ymajorgrids,
        ]
        \addplot+[blue,mark options={fill=blue}] table [x=sentences, y=sentences only, col sep=comma] {data/library.csv};
        \addplot+[red,mark options={fill=red}] table [x=sentences, y=sentences + query, col sep=comma] {data/library.csv};

        \addplot [name path=l sentences query, fill=none, draw=none] table [
            x=sentences, y expr=\thisrow{sentences + query} - \thisrow{CI95 - sentences + query}, col sep=comma]{data/library.csv};
        \addplot [name path=u sentences query, fill=none, draw=none] table [
            x=sentences, y expr=\thisrow{sentences + query} + \thisrow{CI95 - sentences + query}, col sep=comma]{data/library.csv};
        \addplot[red!40, opacity=0.2] fill between[of=l sentences query and u sentences query];

        %\addplot table [x=test_input_size, y=16, col sep=comma] {data/extractive-reader-batch-sizes.csv};
        %\addplot+[mark=triangle*] table [x=test_input_size, y=32, col sep=comma] {data/extractive-reader-batch-sizes.csv};
        %\addplot table [x=test_input_size, y=64, col sep=comma] {data/extractive-reader-batch-sizes.csv};
        %\addplot+[color=orange,mark options={fill=orange}] table [x=test_input_size, y=128, col sep=comma] {data/extractive-reader-batch-sizes.csv};

        %\addplot[dashed,thick, samples=50, smooth,domain=0:6,black] coordinates {(43,0.4)(43,0.9)};
        \addlegendentry{sentences only}
        \addlegendentry{sentences + query}
    \end{axis}%
\end{tikzpicture}%
    \caption{Results for Library test set for various context sizes. The light red area symbolizes a confidence interval with a confidence level of 0.95.  Each point is average from three runs.}
    \label{fig:library_results}
\end{figure}
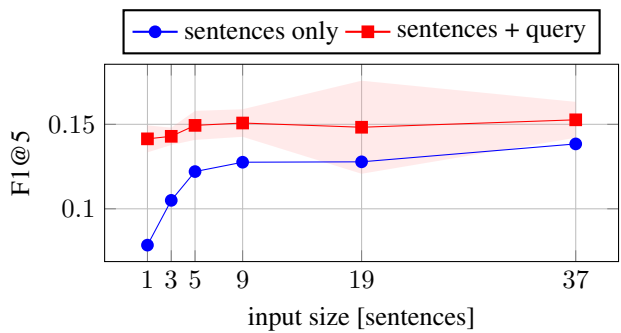

The second set of experiments was performed on our Library dataset. The results can be seen in Figure \ref{fig:library_results}. We have chosen F1@5 because only approximately half of the documents have ten and more keyphrases. Again, the results show that queries are beneficial. Also, it can be seen that the shape of the query curve is similar to Unstructured-SemEval-2010. The average absolute difference between the version with and without query is now 3.1\%. For F1@10, it is 2.3, which is close to the value for the unstructured version of SemEval.

The last set of experiments is done on the Inspec dataset, which has only titles and abstracts. The purpose is to investigate the influence of a query on short inputs containing mainly salient sentences. Results are summarized in Table \ref{tab:results_for_comparison}, which also compares our results with other works. It shows that the results for a single sentence, a single sentence with a title, and whole abstract with a title are similar. The explanation can be that the abstract contains mainly salient sentences containing keyphrases, and also, the abstract itself defines the topic of the article. A similar observation is presented in \cite{kb-rank}, where the version without summarization gives similar results as the extraction performed on a summary.

\begin{table}
\centering
\begin{tabular}{cccc} 
\toprule
\multicolumn{2}{l}{\multirow{2}{*}{}}                                                           & \textbf{Inspec}                    & \textbf{\textbf{SemEval 2010}}  \\ 

\multicolumn{2}{l}{}                                                                            & \multicolumn{1}{r}{\textbf{F1@10}} & \textbf{F1@10}                  \\ 
\midrule\midrule
\multicolumn{2}{r}{TextRank}                                                       & 15.28                     & 6.55                           \\
\multicolumn{2}{r}{KFBS + BK-Rank }                                                       & \textbf{46.62}                     & 15.59                           \\
\multicolumn{2}{r}{DistilRoBERTa + TF-IDF }                                               & -                                  & 16.2                            \\
\midrule
\begin{tabular}[c]{@{}c@{}}\textbf{context}\\\textbf{[sentences]}\end{tabular} & \textbf{query} &                                    &                                 \\ 
\midrule
whole document                                                                 & \ding{55}               & 39.67                              & -                               \\

\multirow{2}{*}{1}                                                             & \ding{55}               & 40.26                              & 14.96                           \\
                                                                               & \ding{51}              & 39.95                              & 16.06                           \\
% \multirow{2}{*}{9}                                                             & \ding{55}               & -                                  & 16.54                           \\
%                                                                               & \ding{51}              & -                                  & 18.49                           \\
\multirow{2}{*}{19}                                                            & \ding{55}               & -                                  & 17.18                           \\
                                                                               & \ding{51}              & -                                  & \textbf{18.56}                  \\
\bottomrule
\end{tabular}
\caption{Comparison of achieved results with other work. KFBS + BK-Rank and TextRank is from \cite{kb-rank}. The DistilRoBERTa + TF-IDF is from \cite{kontoulis-etal-2021-keyphrase}. Our results are averages from five runs.}
\label{tab:results_for_comparison}
\end{table}

\section{Conclusions}
We have conducted experiments that show that query-based keyphrase extraction is promising for processing long documents. Our experiments show the relationship between the context size and the performance of the BERT-based keyphrase extractor. The developed model was evaluated on four datasets; one of them is non-English. The datasets allowed us to find when the query-based approach is beneficial and when not. It was shown that a query gives no benefit when extracting keyphrases from abstracts. On the other hand, it is beneficial for long documents, particularly those without a well-formed document structure on short context sizes.

\section{Acknowledgment}
This work was supported by the Technology Agency of the Czech Republic,  Grant FW03010656 -- MASAPI: Multilingual assistant for searching, analysing and processing information and decision support. The computation used the infrastructure supported by the Ministry of Education, Youth and Sports of the Czech Republic through the e-INFRA CZ (ID:90140).

{
\fontsize{9.5pt}{10.5pt} \selectfont %\small
\bibliography{aaai22.bib}
\bibliographystyle{flairs}
}
\end{document}